\pdfoutput=1

\documentclass[11pt]{article}

\usepackage{acl}

\usepackage{graphicx}
\usepackage{times}
\usepackage{latexsym}

\usepackage[T1]{fontenc}

\usepackage[utf8]{inputenc}

\usepackage{microtype}

\usepackage{tcolorbox}

\newtcbox{\myovalbox}[1][red]{on line,
arc=7pt,colback=#1!10!white,colframe=#1!50!black,
before upper={\rule[-3pt]{0pt}{10pt}},boxrule=1pt,
boxsep=0pt,left=6pt,right=6pt,top=2pt,bottom=2pt}

\title{Extracting Mathematical Concepts from Text}

\author{Jacob Collard \\
  National Institute of Standards and Technology \\
  \texttt{jacob.collard@nist.gov} \\\And
  Valeria de Paiva \\
  Topos Institute \\
  \texttt{valeria@topos.institute} \\\AND
  Brendan Fong \\
  Topos Institute \\
  \texttt{brendan@topos.institute} \\\And
  Eswaran Subrahmanian \\
  National Institute of Standards and Technology \\
  \texttt{eswaran.subrahmanian@nist.gov} \\
}

\begin{document}
\maketitle
\begin{abstract}
  We investigate different systems for extracting mathematical entities from English texts in the mathematical field of category theory as a first step for constructing a mathematical knowledge graph.
  We consider four different term extractors and compare their results. 
  This small experiment showcases some of the issues with the construction and evaluation of terms extracted from noisy domain text.
  We also make available two open corpora in research mathematics, in particular in category theory: a small corpus of 755 abstracts from the journal \emph{TAC} (3188 sentences), and a larger corpus from the nLab community wiki (15,000 sentences).\footnote{Certain commercial entities, equipment, or materials may be identified in this
 document in order to describe an experimental procedure or concept adequately.
Such identification is not intended to imply recommendation or endorsement by the
National Institute of Standards and Technology, nor is it intended to imply that the
entities, materials, or equipment are necessarily the best available for the purpose.}
\end{abstract}

\section{Introduction}
\label{sec:introduction}

The majority of scientific research is communicated using natural language, often in the form of papers like this one. 
However, the volume of scientific literature in any given field is too large to be completely understood by any one individual.
So how can expert researchers, let alone newcomers or outsiders, come to terms with the breadth of scientific knowledge in their field?

Recently, NLP tools have become stunningly effective at making information that is relevant to everyday concerns more accessible. 
Tools for search, question answering, and summarization have improved significantly on various general benchmarks. 
To make research more effective and accessible, similar tools are needed for specialized domains.
Some research communities might number only in the thousands of researchers, and have specialized vocabulary and language usage, including heavy use of symbols, diagrams and/or markup language, as in mathematics.
These smaller communities require a general methodology for constructing specialized tools themselves.

\begin{figure}
    \begin{quote}
    We define the notion of a \myovalbox{torsor} for 
    an \myovalbox{inverse semigroup}, which is based 
    on \myovalbox{semigroup actions}, and prove that 
    this is precisely the structure 
    classified by the \myovalbox{topos} associated 
    with an \myovalbox{inverse semigroup}. Unlike in 
    the \myovalbox{group} case, not all 
    \myovalbox{set-theoretic torsors} are 
    \myovalbox{isomorphic}: we shall give 
    a complete description of the 
    \myovalbox{category of torsors}...
    \end{quote}
    \caption{An example of extracting terms from a single paragraph of text.}
    \label{fig:my_label}
\end{figure}

Knowledge graphs---networks of concepts and their relations in a particular domain of knowledge---have become the preferred technology for representing, sharing, and adding knowledge to modern AI applications~\citep{ilievski2020}. 
The construction of such a graph begins with the identification of central concepts in the domain in question.
Given a corpus of text, such as a collection of papers, the task of identifying these central concepts is sometimes known as \textit{term extraction}, and there are many generic toolkits for performing this task. In this paper we study four examples: TextRank~\citep{mihalcea2004}, DyGIE++~\citep{Wadden2019EntityRA}, OpenTapioca~\citep{delpeuch2020opentapioca}, and Parmenides~\citep{rr2018}.

A potential methodology to construct specialized, domain-specific knowledge management tools would begin by running a generic term extractor over a suitable corpus of domain-specific text and assuming that it extracts a reliable set of terms.
However, each research community may wish to evaluate these terms to test whether they meet the community's specific needs.
This evaluation must determine how well the underlying terms reflect important concepts in the domain.
Ideally, such an evaluation would be made against a corpus annotated by human experts, which would provide a gold standard reference for a representative sample of the domain. 
Such a corpus would ideally capture all and only the relevant concepts present in the corpus, allowing evaluation based on both precision and recall. 

However, obtaining a hand-annotated reference corpus is not always practical, especially with noisy data.
First, hand annotation is time-consuming, and may be infeasible for certain research communities. 
Second, the specialized nature of the text means that the annotators will need to be experts in the domain. 
This makes hand annotation potentially very expensive for highly specialized domains.
In particular, we are also seeking a methodology that can be undertaken with little to no additional direct effort from domain experts, and hand annotation does not meet this criterion.

What, then, does a methodology for constructing and evaluating extracted terms look like for specialized research domains?
In this paper we propose an evaluation methodology that combines information from different `silver standard' sources.
In our case, we study author-selected keywords from paper abstracts, titles from a community-managed wiki, and linguistically identified noun phrases. We argue that, in the case that traditional $F_1$ scores are not informative enough when drawn from any individual source, the evaluation of several sources nonetheless gives us valuable information about the properties of terms extracted.

We apply this methodology to evaluate lists of terms extracted from text in the mathematical field of category theory.
By analyzing the results, we see that generic tools do not have their full efficacy on the specialized domain of category theory, and we have the grounds to infer some reasons why.
Nonetheless, this amalgamated evaluation method provides a path forward for constructing and maintaining a high quality list of domain-specific concepts in category theory.
A key result of this paper is also the groundwork we lay, including two small corpora and some basic experiments, for understanding how NLP tools can be used to build a knowledge graph for mathematics.

\subsection{Related work}
\label{subsec:related-work}

Automatic terminology extraction (ATE) is a well-studied task in natural language process that involves the extraction of domain-specific phrases from a corpus. 
ATE is somewhat distinct from key phrase extraction, which operates at the document level, though the two tasks have some similarities \citep{zhang2018}. 
ATE algorithms often rely on two distinct levels: the identification of linguistic units and the ranking of those units to identify the most relevant and distinctive terms. 
Some algorithms instead identify terms directly, though this usually requires training on an annotated dataset where relevant terms are explicitly identified \citep{Wadden2019EntityRA}.
Work on ATE has been done using large corpora, such as the CiteSeerX library containing millions of scientific documents from many disciplines \citep{patel2020}. 
However, we are not aware of any specific work on ATE for mathematics. 

We are aware of two ACL-style competitions related to mathematical text processing. Firstly, the Math Tasks in NTCIR-10, 11, and 12 studied the recognition of mathematical formulas \citep{Aizawa2021}\footnote{\url{https://ntcir-math.nii.ac.jp/}}.
The second competition is the 2017 SemEval Task 10~\footnote{\url{https://alt.qcri.org/semeval2017/task10/}}, described in \citet{augenstein2017}. 
This task was about extracting keyphrases and relations between them from scientific documents:
the domains chosen were computer science, material science and physics\footnote{\url{https://scienceie.github.io/resources.html}}. Though mathematics itself was not included, all of these disciplines rely on mathematics.

There has also been a great deal of work on technical language processing that is not related to mathematics and does not explicitly involve ATE. 
For example, \citet{Olivetti2020} reviews the use of NLP for materials science, while \citet{perera2020named} covers biomedical information extraction.
The latter is of particular interest due to their use of named entity recognition (NER), which bears some similarity to ATE, and the problems they discuss with recognizing specialized terms.
Generalized approaches face challenges in these domains; as a result, these pipelines make use of domain-specific knowledge bases or expert annotations.

\section{Category theory as a case study}
\label{sec:category-theory}

Although we seek to develop a generic methodology, we have chosen to ground these investigations in the specific field of category theory. 
Category theory is a branch of mathematics focused on relationships and composition. It is often seen as a way to organize mathematics as a whole~\citep{sep-category-theory}.
While this choice is largely dictated by the interests of the authors, category theory presents a number of features which reflect the challenges and potential of automatically constructing domain-specific knowledge management tools. 

Category theory as a  field dates back to the 1940s~\citep{Eilenberg1945}. 
While the field is well established, the volume of text available remains small compared to the corpora used in other NLP applications.
A leading journal in the field, \emph{Theory and Applications of Categories} (TAC), published 55 papers in 2021, and a total of 845 papers since its first issue in 1995. %
This is small compared to, for example, the 3.27 million materials science abstracts used to train the NLP backend for the materials science search engine MatScholar~\citep{kim2017materials}.

Most of category theory research is described in natural language, especially English. 
However, the language is specialized in ways that may pose challenges to automatic systems:

\begin{itemize}
    \item Many technical terms in CT redefine common English words. For example, `category', `limit', `group', `object', and `natural transformation' all have more specific, formalized meanings in CT that they do not have in everyday English.
    \item Many technical terms involve vocabulary that is not present in everyday English at all, such as `groupoid', `monoidal', and `colimit'. 
    \item Special symbols and even diagrams are often interspersed with text, such as `Let $\mathcal{C}$ be a category\ldots'. Often, \LaTeX{} markup is used, and sometimes inconsistent.
    \item Abbreviations and shortcuts are used which would not be common in everyday text, such as the use of `(co)homology' to refer simultaneously to both homology and cohomology.
\end{itemize}

Though the category theory community is relatively small, it has a large online presence, which has supported the creation of community-oriented websites and blogs, including the \emph{n}Lab, a wiki for notes, expositions, and collaborative work, with a focus on category theory.
The \emph{n}Lab was started in 2008, and as of May 2022, has over 16000 articles. 

The authors' own interest and expertise in category theory also allows us to quickly analyze the results of any experiments from the perspective of a potential user.

\section{Automatic Term Extraction Algorithms}
\label{sec:ate-algorithms}

We run a number of experiments to test four different automatic terminology extraction methods: OpenTapioca (a simple entity linking system designed specifically for category theory), DyGIE++ (a neural NER system that has been trained to extract scientific terms), TextRank (a graph-based algorithm originally designed for key phrase extraction, but adapted to ATE), and Parmenides (a linguistically-motivated phrase extraction system that combines symbolic processing and neural parsing).

\paragraph{OpenTapioca:}
OpenTapioca \cite{delpeuch2020opentapioca} is a simple named entity linking system that links phrases of natural language text to entities in WikiData \citep{wikidata}. %
It cannot identify new concepts---only those already represented in WikiData.
OpenTapioca is a simple baseline system that uses basic string matching to identify relevant phrases, built on the recognition that powerful knowledge bases like WikiData has led to recent success in other systems.

OpenTapioca is of particular interest, because it is designed to link entities that are not just locations, dates, or the names of people and organizations, but a variety of technical concepts. 
OpenTapioca also provides a filter that allows it to limit results to entities that appear in \emph{n}Lab, effectively filtering out concepts that are not related to category theory.

\paragraph{DyGIE++:} DyGIE++ \cite{Wadden2019EntityRA} is a span-based neural scientific entity extractor. 
The system builds upon the older DyGIE \cite{Luan2019}.
Both systems were developed in collaboration with the Allen Institute for Artificial Intelligence, and use supervised methods to identify relevant spans of text. 
DyGIE++ has been trained on a variety of different corpora and subtasks, including the identification of chemical compounds, drug names, and mechanisms. 
Though DyGIE++ has not been trained or tested directly on category theory, the similarities between the domains it has been trained on and CT, as well as its overall strong performance, make it a good candidate to test for extracting CT concepts. 

\paragraph{TextRank:}
TextRank \citep{mihalcea2004} is a graph-based ranking algorithm based on PageRank, which has been applied to keyword extraction and text summarization as well as automatic terminology extraction. Though TextRank is a somewhat older algorithm, it is still a common algorithm that has been implemented many times. We use a modern Python implementation, PyTextRank\footnote{\url{https://pypi.org/project/pytextrank/}}.

\paragraph{Parmenides:}
Parmenides~\citep{rr2018} takes a linguistic approach to terminology extraction. 
It uses spaCy\footnote{\url{https://spacy.io}} to identify syntactic structures, then normalizes the syntactic structure and identifies phrases for extraction. 
Parmenides is highly customizable, but is designed primarily for linguistic analysis and not for terminology extraction. 
Nevertheless, it can be used to identify key linguistic phrases as an initial step for ATE.

\section{Test Corpus}
\label{sec:test-corpus}

Automatic terminology extraction takes a corpus of natural language text and produces a list of relevant terms. 
To produce a list of terms for category theory, we need to supply a corpus of category theory text.

To create such a corpus, we take abstracts from \emph{Theory and Applications of Categories} (TAC). 
This is the primary corpus that we use for our experiments.
We also provide a second corpus, using a subset of the \emph{n}Lab wiki\footnote{https://ncatlab.org/nlab/show/HomePage}. 
These corpora will be made publicly available.
We remove markup, section headings, and \LaTeX{} expressions from the text to create a cleaned version of the corpus.
Both corpora are written in English.

After cleaning the corpora, we run spaCy to produce automatic annotations in the style of CoNLL-U. SpaCy is a free open-source library for natural language processing in Python distributed since 2015.
It features named entity recognition (NER), part-of-speech (POS) tagging, dependency parsing, and word vectors.

Note that these are the first publicly available category theory corpora, and we are not aware of any other cleaned, open-source corpora of mathematics research text. 

\begin{table*}[ht]
    \centering
    \begin{tabular}{p{0.4\textwidth}|p{0.55\textwidth}}
    \hline
    \textbf{Reference List} & \textbf{Properties} \\
    \hline
    Author-selected keywords & High precision on advanced, new concepts; poor recall \\
    Page titles from community wiki & High precision on basic concepts; poor recall \\
    Automatically extracted noun phrases & High recall on noun phrases; low precision 
    \end{tabular}
    \caption{The different reference lists under consideration and their properties}
    \label{tab:reference-lists}
\end{table*}

\section{Evaluation Methodology}
\label{sec:evaluation-methodology}

The ATE systems described in Section \ref{sec:ate-algorithms}, combined with the TAC corpus described in \ref{sec:test-corpus}, allow us to construct candidate lists of category theory concepts, which could be used as the basis for a knowledge graph. 
We now arrive at the central question of this paper: how do we assess the quality of such lists?

Again, our goal is not to assess the quality of the extraction algorithms as generic tools, but rather to assess the quality of the lists of category theory concepts they produce.
This is a key distinction: our goal is not generality, but the evaluation of data in a particular context.

More precisely, the usual methodology \citep{Chuang2012WithoutTC} would be to construct an expertly annotated corpus, labeling all the category theory concepts contained within it.
We could then compare the list of terms produced by the term extractors against the gold standard, to produce standard metrics such as precision, recall, and $F_1$ score. 
As described above, this methodology can be expensive and impractical for small, highly technical research communities.

Instead, we seek to evaluate against multiple, imperfect sources of truth to discern different properties of the data.
To compensate for the imperfect nature of our reference lists, we must pair each one with a qualitative description of the properties it can reveal. 
This allows us to use the list to shed light on the nature of the concepts under evaluation, even if a single, representative score cannot be constructed.

The reference lists we consider for this paper are described in Table \ref{tab:reference-lists}.
The properties of each reference list are determined based on how the reference list was constructed.
Author-selected keywords are constructed by human experts to capture the most important concepts in a given abstract.
As a result, they have high precision: all of these elements will be concepts from the field of category theory.
However, they have relatively low recall, because the authors have no incentive to include \emph{all} possible concepts, only the concepts which are new, advanced, or distinctive. 
Thus, many simpler or more common concepts will be excluded from this list.
The page titles from the community wiki, in this case \emph{n}Lab, are similar: they are generally chosen by experts, but will not cover every possible concept.
In this case, basic, common concepts will be covered, but more advanced concepts will not.
Finally, we extract a list of noun phrases uses a pre-trained spaCy model.
This operates under the assumption that many technical terms are noun phrases \cite{Chuang2012WithoutTC}.
This will capture many of these technical terms, but will also capture phrases that are not necessarily technical terms or are only meaningful in context, such as `key results' or `the aforementioned category'. 

While each of these reference lists can give insight on its own, the intersection or union of two or more reference lists can also reveal properties of the extracted terms.
For example, concepts that appear in \emph{both} author keywords and wiki page titles can be understood to be central concepts in the field, so for knowledge graphs, we should focus on having high recall in this area. 
Choosing these reference lists well (i.e., such that their evaluation properties are balanced across desirable properties of our knowledge graph), means we can discover strengths and weaknesses of our extracted term lists.

A key feature of the reference lists that we have chosen is that they incorporate community-maintained, evolving sources. This means that our methodology will be able to improve with increased community effort.
This empowers researchers in the domain to take simple actions that will improve the quality of our term extraction system and its evaluations.

Because the terminology extraction algorithms that we use are all extractive, our reference lists have to be extractive as well.
To ensure this, we filter the phrases in each reference list by comparing them to the TAC corpus.
First, we normalize the phrases using spaCy to remove variations such as morphological inflections and the presence of stop words. 
This allows us to compare terms in the reference list to strings in the corpus to determine if each is present, and remove the terms that are not found in the corpus.

Given an extractive reference list $R$, our evaluation process is fairly standard. For each term extractor $E$ describe above, we:

\begin{enumerate}
    \item Run term extractor $E$ on corpus $C$.
    \item Normalize results using spaCy to get predication list $P$
    \item Produce lists of true positives (appears in both $P$ and $R$), false positive (appears in $P$ but not $R$), and false negatives (appears in $R$ but not $P$).
    \item Calculate recall, precision, and F1 scores.
\end{enumerate}

Note that this produces scores for each reference list, and there is no generic score that covers the extractor in the general case.

\section{Reference Lists}
\label{sec:reference-lists}

\begin{figure*}[ht]
    \centering
    \includegraphics[width=\textwidth]{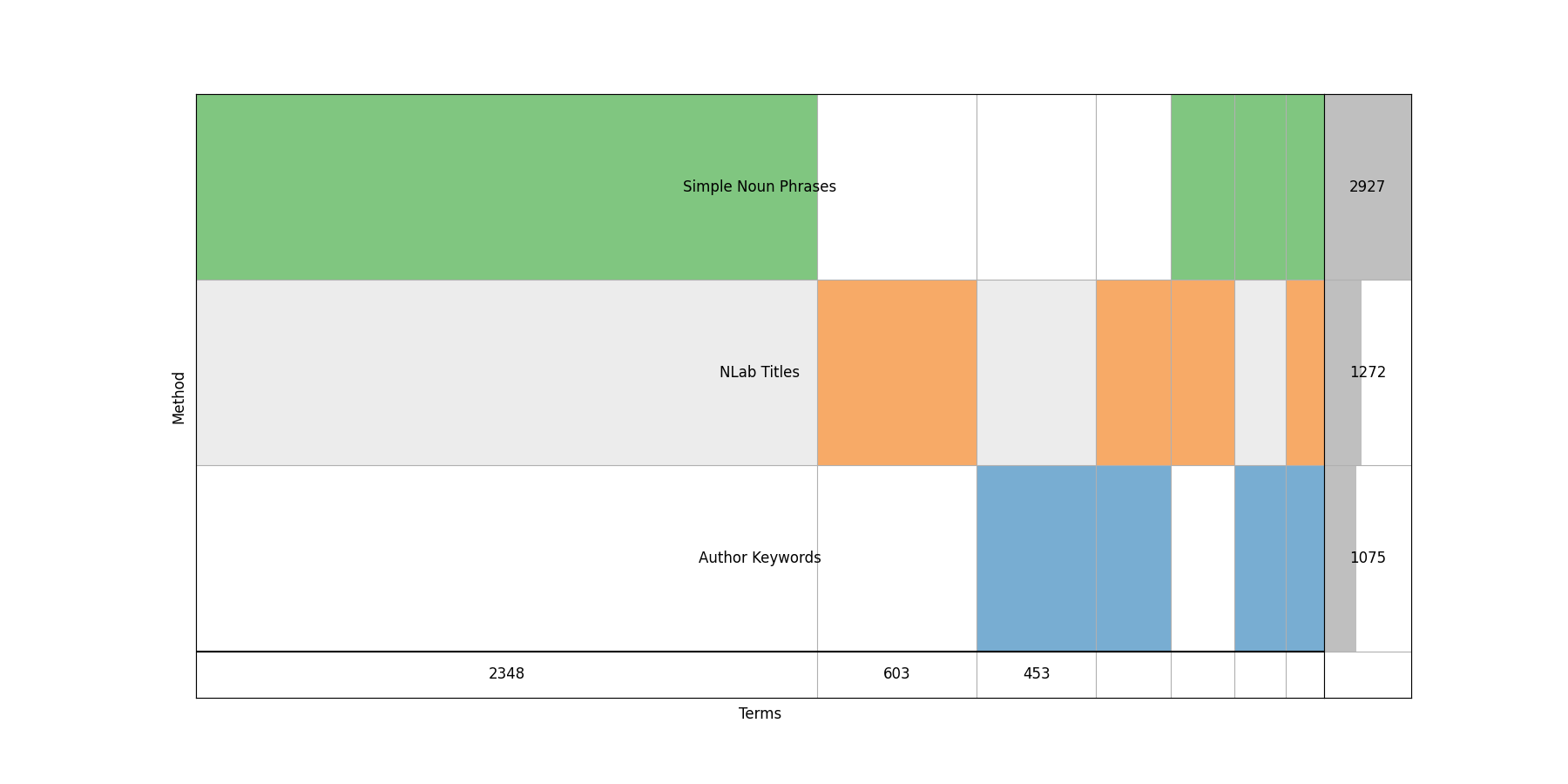}
    \caption{Unique and shared keywords identified by our three reference standards. Each column represents a set of terms; the filled portions of each row represent that the given set of terms was identified by a particular method. For example, the leftmost column shows that 2348 terms were identified only by simple noun phrases. The fourth column shows terms that were identified by both \emph{n}Lab titles and author keywords.}
    \label{fig:standards}
\end{figure*}

We now discuss in more detail the properties of the reference lists we have chosen for category theory. 
Figure \ref{fig:standards} shows the overlap of terms found between the three reference lists.

\subsection{Author Keywords}
\label{subsec:author-keywords}

Our first reference list contains keywords selected by the authors of articles in the journal TAC.

Authors are experts on their own papers.
Author-selected keywords are thus an important, reliable source of truth describing concepts in papers.
However, this reference list has a few complications.
For example, many of the author-selected keywords never show up in the text as described---they are not always \emph{extractive}, and may be more abstract than the terms actually used in the text.
For example, the phrase `topological quantum field theory' could describe the topic of an abstract, but due to its generality, does not necessarily appear in the abstract.
In addition, keywords may contain shortcuts and abbreviations that are easily understood by humans, but not by machines. 
For example, `(co)homology' may be used to describe an abstract that is about both `homology' and `cohomology'. 
Though the normalization described above accounts for author-selected keywords that never show up in texts, it may filter out relevant terms in some cases, such as the `(co)homology' example above, which won't be recognized due to the unusual formatting.
However, this reference list's property of high precision should be maintained due to the authors' expertise.

One final note is that the author keywords are abstract-specific, while ATE is concerned about the corpus as a whole.
Author-selected keywords are still concepts in category theory, but this fact contributes to the lower precision of this reference list: the authors will only select concepts that distinguish their articles from others, and not all concepts that they make reference to.

\subsection{nLab page titles}
\label{subsec:nlab-titles}

Our second reference list is made using page titles from the \emph{n}Lab, a community wiki for mathematics.

In the ideal case, an encyclopedic community wiki would have an article describing every concept in the field.
In practice, this is not the case.
First, the wiki may be initially incomplete, and as the field advances, will lag behind changes in the field.
Second, there may be pages in the wiki that do not necessarily describe concepts \emph{per se}: titles of books, meta-pages, historical notes, and lists do not necessarily belong in a knowledge graph. 
Since we make each reference list extractive, this should not be a significant problem.

This reference list is also very precise, but focuses on concepts that are more likely to be fundamental in category theory, as opposed to more advanced or less common concepts. 
This complements the author keywords well, and shows how well a list of extracted keywords reflects basic concepts in category theory.

\subsection{Noun phrases}
\label{subsec:noun-phrases}

Our third reference list consists of a noun-noun compounds and adjective-noun phrases extracted from the text by spaCy. 
These are all two-word phrases as identified by spaCy's part-of-speech tagger, with \LaTeX{} markup automatically removed.

There is a considerable difference between this reference list and the other two. 
Author keywords and wiki articles are both constructed by experts, and thus clearly belong to the field of category theory.
By contrast, automatically-identified noun phrases, even those taken directly from category theory articles, may not necessarily be mathematical concepts.

\citet{Chuang2012WithoutTC} suggests that around 9.04\% of all keywords chosen by humans are compounds, so this reference list may identify new concepts that are not picked up by other reference lists, though it certainly contains invalid terms, such as `future work' and `next section', as well.

\begin{table*}[ht]
    \centering
    \begin{tabular}{|l|r|r|r|r|}
\hline 
Metric & DyGIE++ & OpenTapioca & Parmenides & TextRank \\
\hline
True Positives & 391 & 236 & \textbf{979} & 600 \\
False Positives & 1105 & \textbf{522} & 13710 & 3231 \\
False Negatives & 684 & 839 & \textbf{96} & 475 \\
Precision & 0.26 & \textbf{0.31} & 0.07 & 0.16 \\
Recall & 0.36 & 0.22 & \textbf{0.91} & 0.56 \\
$F_1$ & \textbf{0.30} & 0.26 & 0.12 & 0.24 \\
\hline
\end{tabular}
    \caption{Extracted terminology compared to author-selected keywords}
    \label{tab:author-keyword-results}
\end{table*}

\begin{table*}[ht]
    \centering
    \begin{tabular}{|l|r|r|r|r|}
\hline 
Metric & DyGIE++ & OpenTapioca & Parmenides & TextRank \\
\hline
True Positives & 399 & 507 & \textbf{1160} & 684 \\
False Positives & 1097 & \textbf{251} & 13529 & 3147 \\
False Negatives & 873 & 765 & \textbf{112} & 588 \\
Precision & 0.27 & \textbf{0.67} & 0.08 & 0.18 \\
Recall & 0.31 & 0.40 & \textbf{0.91} & 0.54 \\
$F_1$ & 0.29 & \textbf{0.50} & 0.15 & 0.27 \\
\hline
\end{tabular}
    \caption{Extracted terminology compared to \emph{n}Lab page titles}
    \label{tab:nlab-results}
\end{table*}

\begin{table*}[ht]
    \centering
    \begin{tabular}{|l|r|r|r|r|}
\hline 
Metric & DyGIE++ & OpenTapioca & Parmenides & TextRank \\
\hline
True Positives & 378 & 216 & \textbf{2439} & 976 \\
False Positives & 1118 & \textbf{542} & 12250 & 2855 \\
False Negatives & 2549 & 2711 & \textbf{488} & 1951 \\
Precision & 0.25 & \textbf{0.28} & 0.17 & 0.25 \\
Recall & 0.13 & 0.07 & \textbf{0.83} & 0.33 \\
$F_1$ & 0.17 & 0.12 & 0.28 & \textbf{0.29} \\
\hline
\end{tabular}
    \caption{Extracted terminology compared to noun phrases}
    \label{tab:noun-phrase-results}
\end{table*}

\begin{table*}[ht]
    \centering
    \begin{tabular}{|l|r|r|r|r|}
\hline 
Metric & DyGIE++ & OpenTapioca & Parmenides & TextRank \\
\hline
True Positives & 748 & 547 & \textbf{3606} & 1653 \\
False Positives & 748 & \textbf{211} & 11083 & 2178 \\
False Negatives & 3518 & 3719 & \textbf{660} & 2613 \\
Precision & 0.50 & \textbf{0.72} & 0.25 & 0.43 \\
Recall & 0.18 & 0.13 & \textbf{0.85} & 0.39 \\
$F_1$ & 0.26 & 0.22 & 0.38 & \textbf{0.41} \\
\hline
\end{tabular}
    \caption{Extracted terminology compared to the combined reference lists}
    \label{tab:combined-results}
\end{table*}

\section{Results}
\label{sec:results}

Analyses of each corpus, with respect to all three reference lists, can be found in our GitHub repo. 
Summaries of the results of our experiments are given in Tables \ref{tab:author-keyword-results}, \ref{tab:nlab-results}, and \ref{tab:noun-phrase-results}.
We also evaluate the results against the union of all three reference lists, as shown in Table \ref{tab:combined-results}.

Further results are described in the supplementary data, including the results of earlier experiments. Overall, however, the general ranking of the term lists remains the same, with few exceptions. 

\section{Discussion}
\label{sec:discussion}

Overall, the $F_1$ scores presented here are very low when compared to the results of SEMEVAL 2017 \citep{augenstein2017}.
DyGIE++ also reports higher numbers on the datasets it has been trained on \citep{Wadden2019EntityRA}.
Our results are, however, similar to the results of \citet{patel2020}, which considers the problems of terminology extraction using papers indexed in CiteSeerX, which reports $F_1$ scores of 0.33. 

Parmenides always outperforms the other models we consider on recall, but generally performs poorly on precision.
Conversely, OpenTapioca has relatively high precision scores, resulting in the highest $F_1$ score for both author keywords and \emph{n}Lab page titles.
Parmenides was designed as a linguistic analysis tool; it extracts all possible phrases, with only limited power to rank those phrases by relevance.
As a result, it extracts almost all of the linguistic units that are available, including large amounts of irrelevant text. 
OpenTapioca, on the other hand, is designed to pull out only category theory concepts, but is limited in its ability to extract novel terms and those not described in \emph{n}Lab.

The terms extracted by DyGIE++ are reasonable in terms of $F_1$ score. 
For author-selected keywords, DyGIE++ performs the best, and it has the second-highest $F_1$ score for \emph{n}Lab page titles.

However, it is not enough to just consider $F_1$ scores in this case. 
The reference lists that we consider have limitations, and we cannot rely on them all to be both complete and precise.
As described above, the author keywords and \emph{n}Lab titles have limited \emph{recall}---they do not contain all of the possible category theory terms in the text, because they are designed for other purposes. 
For these reference lists, we can only rely on the recall of the extracted terms. 
Low recall on the author keywords indicates that a list does not contain many of the advanced concepts from category theory, while low recall on the \emph{n}Lab titles indicates that a list does not contain many of the basic concepts from category theory.
Low precision on these, however, may indicate that a list contains terms which may still be valid, but which do not appear in these reference lists.

The proper conclusion, then, should not be that OpenTapioca is the best option because it has the best overall $F_1$ score.
Nor is DyGIE++ necessarily ideal just because of its high performance on author-selected keywords. 
OpenTapioca, as shown by low recall on noun phrases, cannot extend well to novel terms.
DyGIE++ performs reasonably well overall, but is outperformed by several extractors in recall of \emph{n}Lab page titles and by Parmenides and TextRank on recall of author-selected keywords.
Instead, TextRank appears to be the best candidates, having high recall on author-selected keywords and \emph{n}Lab page titles as well as high precision on noun phrases, though a better measure of precision is desirable.

%
 

\section{Conclusions}
\label{sec:conclusions}

We present the first computational work extracting mathematical concepts from abstracts.
We investigated four different term extractors, previously described for other domains, and evaluated the results against the limited annotated data we had for category theory.
The results are somewhat limited as well, compared to previous results on more generic domains. 
However, other domain-specific analyses have some of the same problems, which suggests that our results are still promising.

We also provide insight into the evaluation of automatically-generated terminologies for limited-resource domains.
The usual $F_1$ scores are not entirely reliable unless the gold standard can be assumed to include both all and only the relevant terms, but partially-correct `silver standards' may still provide useful insight into the data.

In our case, we can draw some important conclusions about the terminology lists that we extract.
Because author keywords and \emph{n}Lab titles are most reliable for recall, we can determine that tools such as Parmenides and TextRank are able to extract large quantities of both advanced and basic category theory terms.
However, the large number of other terms extracted by Parmenides suggests that it may need additional filtering to be useful for automatic terminology extraction for our use-case. 
Our evaluation can also be further improved.
The low relative recall of the noun phrase reference list itself suggests that additional phrase types are common in our data. 
Adding verb phrases and more complex noun phrases could help us identify high-precision terminologies, as well as high-recall ones.

Another possibility in our case is to continue working toward our use-case.
Since we have further downstream uses of the terminology---namely, the creation of a knowledge graph---we can use this to further our evaluation.
By extracting relations between terms, we can identify which terms are the most connected and which are isolated, under the assumption that isolated terms are less likely to be part of domain-specific language.

We have also constructed two publicly-available corpora that can be developed into more sophisticated datasets.
Though there are still many limitations to both evaluation and ATE in mathematics, we hope that our work provides a basis for future developments in the area, and that our insights on evaluation and domain-specific research can be applied more generally.

\bibliography{anthology,custom}
\bibliographystyle{acl_natbib}




\end{document}